\documentclass[preprint,12pt]{elsarticle}

\usepackage{amssymb}
\usepackage{amsmath}
\usepackage{algorithmic}
\usepackage{graphicx}
\usepackage{textcomp}
\usepackage{xcolor}
\usepackage{stfloats}
\usepackage{url}
\usepackage{verbatim}
\usepackage{caption}
\usepackage{float}
\usepackage{subcaption}
\usepackage{array}
\usepackage{booktabs}
\usepackage{multirow}
\usepackage{orcidlink}

\begin{document}

\begin{frontmatter}

\title{Structure-Guided Multi-Granularity Spatiotemporal Transformer for Efficient Monocular 3D Human Pose Estimation}

\author[A]{Sheng Xu}
\author[B]{Aoduo Li}
\author[D]{Rongli He\corref{cor1}}
\cortext[cor1]{Corresponding author: Rongli He}

\affiliation[A]{organization={School of Physics and Optoelectronic Engineering,Guangdong University of Technology},
            city={Guangzhou},
            country={China}}
\affiliation[B]{organization={School of Advanced Manufacturing, Guangdong University of Technology},
            city={JieYang},
            country={China}}
\affiliation[D]{organization={College of Engineering, South China Agricultural University},
            city={Guangzhou},
            country={China}}

\begin{abstract}
In monocular 3D human pose estimation, existing multi-granularity methods focus on spatial joint features but overlook the structured information within skeletal sequences, such as global connectivity and motion continuity. Such structured information is not effectively integrated into the Transformer architecture, hindering collaborative modeling. In addition, full-frame sequence modeling increases computational cost and reduces efficiency. To address these challenges, this paper proposes a Structure-Aware Multi-Granularity Spatiotemporal Transformer for efficient and accurate pose estimation by the guidance of structured information through the alignment and fusion of multi-granularity features. Specifically, a three-level granularity feature extractor is designed to model human topology at the joint, bone, and limb levels. For the  spatial multi-granularity modeling, we propose a Gated Adaptive Spatial Attention (GASA) mechanism that explicitly incorporates structural constraints into the spatial attention computation, enabling precise modeling of spatial structured information. Additionally, a Multi-Granularity Feature Enhancement (MGFE) mechanism is designed to perform hierarchical pooling fusion to strengthen spatial local detail and global semantic representation. For the temporal multi-granularity modeling, a Temporal Segment Fusion (TSF) mechanism is employed to align temporal context, enhancing both local and global temporal modeling capabilities. Finally, a Hierarchical Cross-Attention (HCA) mechanism enables cross-granularity spatiotemporal feature interaction, enhancing structured information fusion and pose estimation accuracy. A Keyframe Attention (KFA) mechanism dynamically selects key motion segments, reducing redundancy and computational cost. Experiments show that the proposed model outperforms conventional Transformer methods in accuracy and efficiency, offering an effective and robust solution for monocular 3D human pose estimation. 
The source code is publicly available at \url{https://github.com/AsakaTigar/3D-Pose-estimation}.
\end{abstract}


\end{frontmatter}

\section{Introduction}
\label{intro}
Three-dimensional human pose estimation (3D HPE) aims to recover the positions of key skeletal points of the human body in three-dimensional space from images or videos, and is widely used in human-computer interaction, behavior recognition, virtual reality, and other fields \cite{martinez2017simple, wang2021review, sun2022survey}. This task is essentially an inverse problem of inferring three-dimensional structure from two-dimensional image information, which is significantly ill-posed and easily affected by factors such as occlusion, depth blurring, and cluttered backgrounds. Therefore, fully leveraging auxiliary clues from temporal sequences, spatial structures, or multi-view perspectives is crucial for improving estimation accuracy.

In practical applications, monocular 3D pose estimation (Monocular 3D HPE) has garnered particular attention. Compared to multi-camera systems, monocular systems are simpler to deploy and more cost-effective, making them more suitable for real-time tasks in open environments \cite{sun2022survey}. However, due to the lack of parallax information, relying solely on a single frame image to recover 3D structure faces the issue of severe information deficiency. Although numerous methods have attempted to introduce temporal consistency or structural constraints \cite{li2021survey} in recent years, it remains challenging to extract both global and local motion semantics from a single granularity, thereby affecting the stability and generalization capability of pose prediction.

To address these issues, researchers have gradually begun to explore multi-granularity modeling mechanisms. By abstracting human body structure at different levels—such as from overall anatomical structure, major trunk joints, to basic motion joints—the system can systematically model topological and semantic information of the human body in both spatial and temporal dimensions. This not only enhances representational capabilities but also better leverages the complementary nature of local details and global context\cite{li2022mstformer}, but the interaction and fusion between features of different granularities are often insufficiently deep, preventing the model from fully leveraging the complementary information between features of different granularities \cite{li2022mstformer, luo2021graformer}. For example, in multi-granularity algorithms based on GCN, the fusion of local joint features and global skeleton features is often merely a simple concatenation or weighted summation which fails to effectively capture the complex relationships between local joint movements and changes in the global skeleton structure \cite{duan2022glagcn}. For instance, during jumping, local joint changes are closely linked to global motion trends, but existing fusion methods fail to capture this, limiting pose estimation robustness. High-precision monocular 3D human pose estimation algorithms often require a large number of frames as input, which to some extent leads to the issue of excessive redundant frames \cite{chen2023sct, li2023kinematics}. However, current multi-granularity algorithms fail to effectively identify and discard redundant frames, treating all frames equally and thereby increasing computational overhead.

The current main methods for monocular 3D human pose estimation can be broadly categorized into the following types: first, image-to-3D regression methods, which directly regress 3D keypoint coordinates from a single RGB image, typically using convolutional neural networks (CNNs) or their variants to establish the mapping relationship between images and 3D poses \cite{martinez2017lifting, martinez2017simple}. Second, 2D-to-3D pose lifting methods adopt a two-stage strategy of “2D first, then 3D.” They first use a mature 2D pose detector to obtain 2D keypoints of the human body, then map them to 3D poses. This approach can leverage large-scale 2D annotated data to improve the accuracy of the front-end detection. However, due to the inherent projection ambiguity of 2D keypoint information (a single 2D pose may correspond to multiple 3D solutions), this method inherently remains an ill-posed problem \cite{zhang2021unsupervised}. Third, temporal modeling methods that incorporate temporal information aim to address the limitations of single-frame image data by utilizing consecutive frames from videos to obtain more stable 3D pose estimates through temporal modeling \cite{pavllo2019, chen2016exploiting}. However, information between video frames is often highly redundant, and modeling with short time windows has limited effectiveness; modeling with long sequences significantly increases computational costs. Fourth, Transformer methods leverage their strong ability to model long-range dependencies and global perception for pose estimation tasks. For example, methods like PoseFormer and LiftFormer use spatio-temporal self-attention modeling on 2D keypoint sequences to achieve more accurate pose recovery \cite{li2022liftformer, zheng2021poseformer}. 

Although Transformers demonstrate strong global modeling capabilities in 3D human pose estimation, they still have many shortcomings in practical applications. First, traditional Transformers mainly rely on unified modeling of all joints and lack a detailed distinction between human structural levels, resulting in insufficient utilization of structural information and difficulty in modeling complex movement structures with precision \cite{zeng2022hstformer}. Second, Transformers typically fuse information of different granularities using simple concatenation or weighting methods, ignoring semantic alignment between granularities, which hinders the collaborative expression of multi-level motion information \cite{li2022mstformer}. Third, Transformers typically process entire video segments uniformly, unable to effectively identify key action frames, leading to redundant information interfering with modeling and reducing computational efficiency \cite{li2022poseformerv2, yuan2023stie}.

To address these issues, this paper proposes a multi-granularity spatio-temporal Transformer network that integrates skeletal structural information, achieving efficient and accurate monocular 3D pose estimation. 
This paper introduces a progressive three-level granularity modeling framework, which hierarchically represents the human body structure at the levels of the full body, torso, and motion joints, enabling fine-grained modeling of human topology. Specifically, to tackle the  insufficient utilization of structured information, a Gated Adaptive Spatial Attention (GASA) mechanism is proposed to explicitly incorporate skeletal connectivity constraints into the attention computation, thereby achieving precise modeling of spatial structural information.
To overcome the lack of semantic alignment in multi-granularity feature fusion, a Multi-Granularity Feature Enhancement Module (MGFEM) is designed into the Spatial Transformer. It performs hierarchical pooling and fusion to integrate global contextual information at each granularity level, enhancing the representation of local details and global semantics. In the Temporal Transformer, a Temporal Segment Fusion (TSF) mechanism is proposed to align the temporal context and improve the modeling of local motion dynamics. Furthermore, a Hierarchical Cross-Attention (HCA) mechanism is designed to enable spatiotemporal semantic alignment and collaborative modeling across granularities.
Finally, to reduce redundancy and improve efficiency in sequence processing, a Keyframe Attention (KFA) mechanism is proposed to dynamically select key motion frames, effectively reduce computational cost while preserving temporal modeling of essential motion cues.

The contributions of this paper are:
\begin{enumerate}
\item Addressing the limitations in structure-aware modeling and multi-granularity semantic alignment in existing Transformers for 3D human pose estimation, we establish a unified framework that integrates structure awareness, multi-granularity alignment, and efficient temporal modeling, thereby improving the accuracy and computational efficiency.
\item We propose a structure-aware Spatial Transformer that adaptively embeds structural information into multi-granularity spatial attention, together with a feature enhancement mechanism. Meanwhile, a Temporal Segment Fusion mechanism is designed in the temporal Transformer to align temporal context.
\item A hierarchical cross-attention mechanism is designed to achieve semantic alignment and fusion of multi-granularity features, overcoming the semantic drift issues associated with traditional fusion.
\item Key Frame Attention is proposed to dynamically select temporal key segments for attention modeling, effectively reduce computational costs while maintaining the ability to model temporal dependencies in long sequences.
\end{enumerate}

\section{Related Work}
\subsection{Multi-granularity Feature Modeling}
The human skeleton naturally has a layered anatomical structure, and joints at different levels have different semantic meanings when expressing movement, posture, and overall dynamics. Therefore, multi-granularity modeling has gradually become an important direction for improving posture estimation performance. The core idea of multi-granularity modeling is to abstract and model the human skeleton at multiple scales and structural levels, enabling the model to capture both local details (such as fine movements of the arms and legs) and understand overall movement semantics (such as the orientation of the torso and the overall dynamic trends of the body).

Current multi-granularity modeling methods can be broadly categorized into three types: first, hierarchical structural modeling methods, which divide the skeleton into multiple semantic layers based on human anatomical structure. For example, SemGCN \cite{yan2019semgcn} uses GCN to perform structural modeling on groups of joints and fuses whole-body information through hierarchical graphs. Second, multi-scale feature extraction methods extract features at different spatial scales to achieve both local and global semantic coverage. For instance, PoseGTAC \cite{shan2022posegtac} designs graph dilated convolutions and graph Transformer modules to extract local fine-grained and global coarse-grained information, respectively. LiftFormer \cite{li2022liftformer} uses multi-scale attention mechanisms to learn cross-temporal granularity structural representations. Third, partition-based modeling and feature enhancement methods divide the human body into functional regions. FLEX \cite{zeng2020flex} fuses 2D joint coordinates and confidence scores, using partition-based modeling to extract stable local features.

Existing methods still have some limitations: for example, the granularity division is fixed, making it difficult to adapt to dynamic changes in action semantics; the interaction mechanisms between granularities are simple, often involving concatenation or weighting, lacking deep interaction modeling; and there is no systematic integration of granularity structure and temporal dimension information, making it difficult to capture the continuity of dynamic structures \cite{sun2022survey}. To address these issues, we achieves effective interaction between granularities through dynamic gated fusion and cross-attention mechanisms.

\subsection{Skeletal Sequence Structural Information}
In 3D human pose estimation, skeletal sequences not only carry the geometric positions of human joints but also implicitly contain rich structural information such as topological connections and motion patterns. Existing methods can be categorized as follows: Graph-based Structural Modeling methods model the human skeleton as a graph structure. ST-GCN \cite{yan2018stgcn} was the first to introduce spatiotemporal graph convolutions. SemGCN \cite{yan2019semgcn} further introduced semantic priors to enhance contextual modeling. PoseGraphNet \cite{ci2020posegraphnet} explores different graph structure designs. Prior-constrained modeling methods explicitly utilize physical constraints. Fang et al. \cite{fang2018grammar} proposed a method based on pose grammar. FLEX \cite{zeng2020flex} enhances structural stability by jointly predicting bone length and angle information. Temporal-structural modeling methods focus on the continuity of joint movements. Pavllo et al. \cite{pavllo2019} proposed a framework based on temporal convolutions. Strided Transformer \cite{li2022strided} introduces a hierarchical Transformer to model long-range temporal dependencies. Cai et al. \cite{cai2019gcn} unify spatial structure and temporal information modeling.

The above methods still have several issues: First, most methods rely on static topology or artificial priors, making it difficult to dynamically adapt to diverse human motion structures; Second, graph structures often model joint connections at a single granularity; Third, spatio-temporal structure modeling is mostly handled separately. To address this, this paper designs a dynamic gated fusion mechanism to explicitly encode structural information into attention calculations and a multi-granularity cross-attention mechanism to achieve alignment across granularities.

\subsection{Spatio-Temporal Transformers in Pose Estimation}
Transformers have been widely adopted in 3D human pose estimation due to their powerful global and sequence modeling capabilities. Current Transformer-based methods can be categorized as follows: First, Temporal Transformer from 2D Keypoints methods focus on using 2D keypoint sequences as input. A typical example is PoseFormer \cite{zheng2021poseformer}. MHFormer \cite{li2022mhformer} introduces a multi-hypothesis modeling mechanism. Second, Spatio-temporal Transformer methods jointly model spatial structures and temporal dynamics. LiftFormer \cite{li2022liftformer} implements an efficient conversion from 2D to 3D pose. Strided Transformer \cite{li2022strided} introduces strided convolution to progressively aggregate temporal information. Third, Multi-view and Temporal Fusion Transformer methods address high redundancy in multi-view and video scenarios. MTF-Transformer \cite{he2022mvg} designs multi-view and temporal fusion modules. TransFusion \cite{zhang2022transfusion} utilizes Transformers to establish cross-frame semantic associations.

Existing methods still have issues: First, most do not fuse structured topological information; Second, the fusion of spatio-temporal information is typically modeled sequentially; Third, models often uniformly process video sequences without a keyframe selection mechanism. To address these issues, our proposed network designs a progressive multi-granularity feature extraction structure, combines feature enhancement and cross-attention mechanisms, introduces a dynamic gating fusion module, and uses a keyframe attention mechanism to balance efficiency and accuracy.

\section{Method}
\subsection{Problem Analysis and System Design}
Although Transformers have been widely used in monocular 3D human pose estimation, existing methods still lack structure-aware modeling, often rely on static priors, and fail to align multi-granularity features across skeletal hierarchies\cite{zhou2017deepkinematics}\cite{li2022mstformer}. Moreover, uniform temporal modeling introduces redundancy and reduces computational efficiency\cite{cibrario2024hourglass, li2022poseformerv2}.

To address the lack of structured information utilization, we introduce GASA to dynamically embed skeletal connection relationships into the attention calculation for multi-granularity, Compared to existing GCNs that rely on fixed adjacency matrices or Transformers that are fully data-driven, GASA can dynamically adapt to skeletal dependencies under varying actions and pose changes. To address the issue of insufficient granularity alignment, We propose MGFE and HCA for better feature alignment. Unlike approaches that use only single-granularity features or simple concatenation, our approach fully leverage the complementary strengths of joints, bones, and limbs in capturing fine details and understanding structural context. To address the inefficiency in temporal modeling, we proposed TSF and KFA to improve efficiency. Unlike conventional methods that uniformly process entire video sequences or rely on fixed-length temporal windows, our approach adaptively aligns temporal context and selectively focuses on key motion frames.
\begin{figure}[htbp]
\centering
\includegraphics[width=\textwidth]{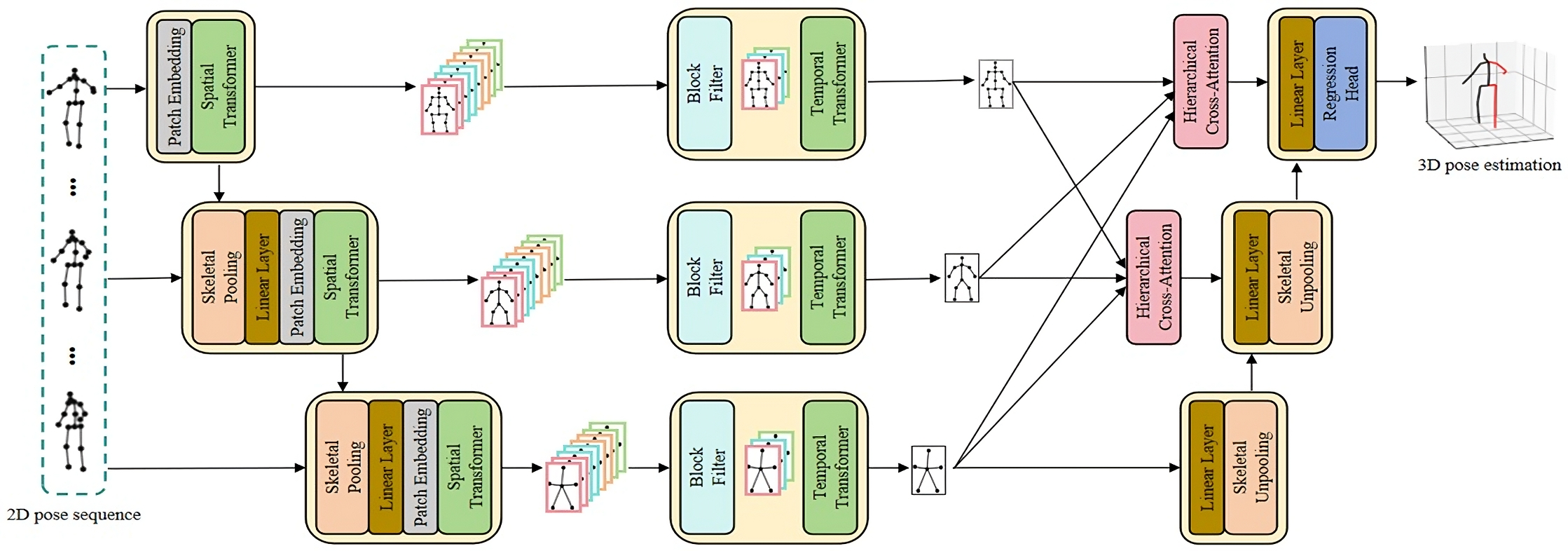}
\caption{The overall architecture of our proposed network (Structure-Aware Aligned Multi-Granularity Spatiotemporal Transformer). The model processes a 2D pose sequence through three parallel, multi-granularity streams. Features are progressively refined and then fused via a Hierarchical Cross-Attention mechanism before being up-sampled for the final 3D pose estimation.}
\label{fig:overall_structure}
\end{figure}

As shown in Fig.~\ref{fig:overall_structure}, the overall process is as follows: first, multi-granularity features are extracted via grouped pooling. Then, spatial and temporal modeling are performed sequentially using our proposed GASA, MGFE, TSF, and KFA mechanisms. Subsequently, HCA is used for feature interaction. Finally, inverse pooling restores fine-grained features for pose estimation.

\subsection{Multi-scale Skeleton Topological Representation}
\begin{figure}[htbp]
\centering
\includegraphics[width=0.7\columnwidth]{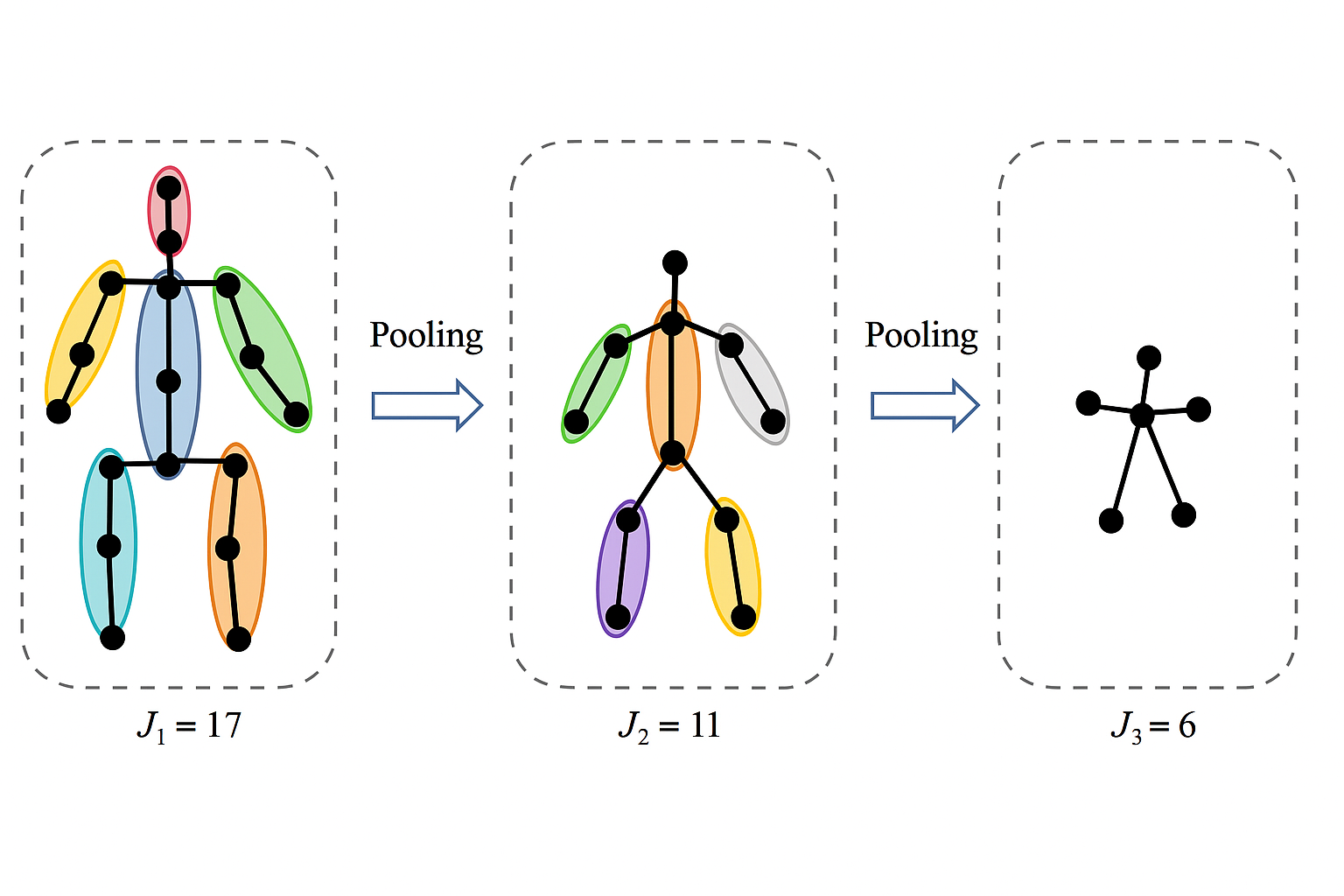}
\caption{Representation of our multi-scale skeletal topology. The skeleton is progressively simplified from 17 joints (J1) to 11 joints (J2) and finally to 6 joints (J3) through a hierarchical pooling strategy.}
\label{fig:topo_rep}
\end{figure}

Based on the anatomical structure of the human skeleton \cite{ci2020posegraphnet}, we divide it into six groups $\left\{X^g \mid g=1,2, \ldots 6\right\}$: head, torso, left arm, right arm, left leg, and right leg. Fig.~\ref{fig:topo_rep} shows our designed three-scale skeleton topology. The number of keypoints at each level is $J_1=17, J_2=11, J_3=6$.

After inputting the monocular feature sequence, the grouped skeleton pooling layer extracts coarse-grained skeleton features from the fine-grained ones using 2D convolution. The group-wise pooling operation is defined as:
\begin{equation}
\hat{X}_i^g = \operatorname{Conv}_{2D}\left(P_{i+1}^g, X_i^g\right)
\end{equation}

where the skeleton is partitioned into groups based on human body parts. The input features of the $(i+1)$-th skeleton pooling layer can be denoted as $\{X^g \mid g = 1, 2, \ldots, 6\}$, where $X_i^g \in \mathbb{R}^{N_i \times C_i}$ represents the feature of group $g$ at level $i$, 
 ${R}^{N_i \times C_i}$ is a feature matrix for group $g$ with $N_i$ joints and $C_i$-dimensional features per joint, and $P_{i+1}^g \in \mathbb{R}^{N_{i+1} \times N_i}$ is the learnable parameter matrix that reduces the number of keypoints in group $g$ from $N_i$ to $N_{i+1}$.

Following the multi-granularity spatiotemporal alignment by HCA at the final stage of the pipeline, The unpooling layer applies 2D transposed convolution to each group individually, restoring coarse-grained skeletal features to fine-grained high-resolution representations $\hat{\mathbf{F}}_{i}^{g}$ for pose estimation:
\begin{equation}
\hat{F}_i^g=\operatorname{ConvT}_{2 D}\left(\left[U_i^g\right]^T, F_i^g\right)
\end{equation}

In the $i$-th skeleton unpooling layer, the input features represented as 
\(\{F_i^g \mid g = 1, 2, \ldots, 6\}\) which passed through the unpooling operation 
to restore the structural resolution. Specifically, 
\([U_i^g]^\top\) is a learnable parameter that upsamples group $g$ from \(N_i\) 
keypoints to \(N_{i-1}\) keypoints.

\subsection{Gated Adaptive Spatial Attention (GASA)}
To integrate skeletal topology priors, we propose a Gated Adaptive Spatial Attention (GASA) mechanism that incorporates multi-scale skeletal connectivity into attention computation to enhance spatial relationship modeling between keypoints. 

We construct a multi-level skeletal adjacency matrix $\mathbf{A}^{(i)} \in \{0,1\}^{J_i \times J_i}$, where $A_{mn}^{(i)}=1$ if joint $m$ and $n$ are physically connected, otherwise $A_{mn}^{(i)}=0$. 
Based on this, we design a dynamic gating mechanism to adaptively fuse prior knowledge and original attention. A gating matrix \( G \) is generated via a fully connected layer followed by a Sigmoid function $G=\sigma\left(f\left(X_0\right)\right) \in[0,1]^{B \times 1 \times 1 \times J_i}$, This gating matrix adaptively fuses the two attention branches to produce the final attention weights:
\begin{equation}
A_{\text{final}} = \text{Softmax}\left( (1 - G) \cdot \hat{A} + G \cdot A^{(i)} \right) \tag{4.1}
\end{equation}

where \(\hat{A}\) denotes the original attention matrix, and \(A^{(i)}\) is the adjacency matrix at layer.  This formulation allows each keypoint to adaptively balance between prior structural knowledge and local attention cues. The final attention matrix \(A_{\text{final}}\) is used to update features by multiplying with the value matrix \(V\).

\subsection{Multi-Granularity Feature Enhancement (MGFE)}
\begin{figure}[htbp]
\centering
\includegraphics[width=0.63\columnwidth]{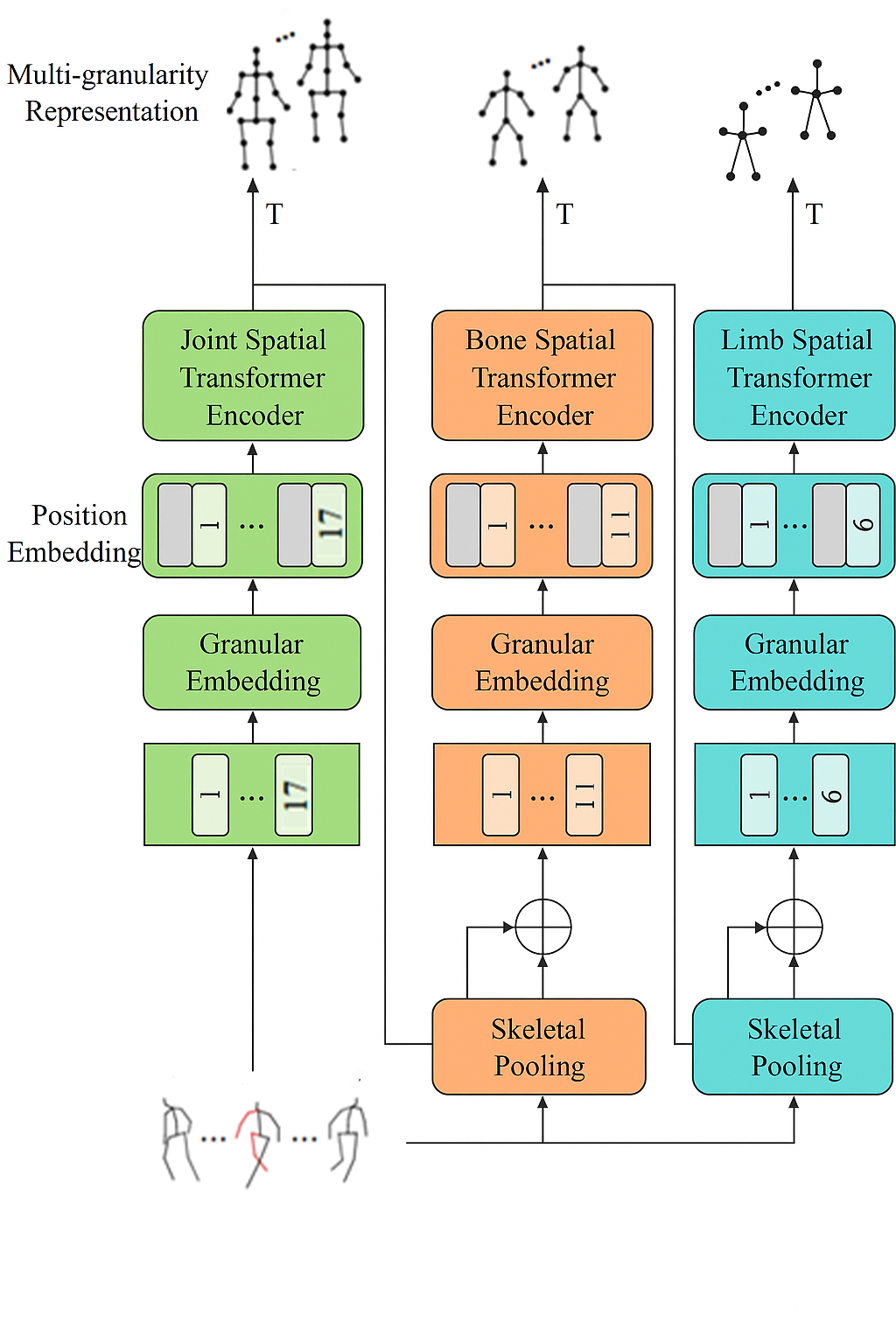}
\caption{Illustration of the multi-granularity feature enhancement mechanism (MGFE). At the bone and limb levels, features from the original input are pooled and concatenated with the output of the previous, finer-grained encoder.}
\label{fig:mgfe}
\end{figure}

To improve the model's ability to extract spatial features, we propose a Multi-Granularity Feature Enhancement (MGFE) mechanism, as shown in Fig.~\ref{fig:mgfe}. From left to right are the first to third levels of feature extraction. For the second and third levels, we fuse the pooled features from the raw input with the output features from the previous level, as shown in Equation (4.2):
$$
F_i=\operatorname{Transformer}\left(F_{i-1} \oplus \operatorname{Pool}\left(F_0\right)\right)
$$

Here, $F_{i-1}$ represents the output features from the $(i-1)$-th level, $\operatorname{Pool}\left(F_0\right)$ denotes the pooled features from the original input $F_0$, and $\oplus$ denotes feature concatenation.

After extracting joint-level spatial correlations via the joint-level Spatial Transformer, to enhance the utilization of raw input information, we apply pooling to the original input and concatenate the resulting features with the joint-level outputs from the previous layer. This combined feature is then fed into the organ-level Spatial Transformer to extract organ-level structural information. At the third level, we again pool the original input, concatenate it with the output from the previous level, and feed it into the limb-level Spatial Transformer to model spatial dependencies at the limb level.

The pooled features from the raw input contain global skeletal structure information, which serves as prior knowledge and enriches the spatial representation. Fusing these global features with local features from the previous level allows the model to capture not only localized spatial dependencies (e.g., joint and bone structures) but also global contextual information. This combination improves the model's understanding of the spatiotemporal nature of the human skeleton across different levels.
For example, in multi-step or compound actions, local features help capture detailed movement patterns, while global features provide broader structural context. Their combination enables the model to more accurately estimate the 3D pose.

\subsection{Temporal Segment Fusion (TSF)}
To enhance temporal feature capture, we introduce a Temporal Segment Fusion (TSF) mechanism. By explicitly introducing temporal context through feature concatenation, the model achieves faster training and improved temporal sequences understanding. Before the temporal Transformer, for each time step $t$, we concatenate the first half of the feature vector from the previous time step ($t-1$) with the second half of the current time step's feature vector:
\begin{equation}
\mathbf{x}'_t = \mathbf{x}_{t-1}^{(1)} \oplus \mathbf{x}_{t}^{(2)}
\label{eq:tsf}
\end{equation}

where $\mathbf{x}^{(1)}$ and $\mathbf{x}^{(2)}$ denote the first and second halves of the feature vector, respectively, and $\oplus$ denotes concatenation. 

The temporal segment fusion mechanism enables the model to explicitly perceive the relationships between adjacent time steps during training. Instead of learning temporal continuity from scratch, the model can directly leverage the concatenated features, thereby accelerating training and improving its ability to model temporal dependencies. 

\subsection{Keyframe Attention Mechanism (KFA)}
We propose a temporal Transformer module based on sparse attention across blocks, called the Keyframe Attention (KFA) mechanism. First, the temporal dimension $T$ is divided into $N$ equal-length sub-blocks of length $r=T/N$ in the previous step, Then, block-level features $X_{\text{block}} \in \mathbb{R}^{B \times P \times C}$ are generated through average pooling. To filter key temporal segments, we use a gating mechanism to compute attention weights for each block:
\begin{equation}
\mathbf{A} = \sigma(\mathbf{W}_{2} \text{ReLU}(\mathbf{W}_{1} \cdot \text{AvgPool}(\mathbf{X}_{block})))
\label{eq:kfa}
\end{equation}

where $\mathbf{X}_{block} \in \mathbb{R}^{B \times N \times P \times C}$ and $\mathbf{W}_1, \mathbf{W}_2$ are trainable parameters. We select the top-$k$ blocks, reducing computational complexity from $O(T^2)$ to $O(k^2)$. The selected blocks are then treated as patches and processed by a standard Transformer encoder with position embeddings \cite{vaswani2017attention}.

The selected sparse blocks are passed through the temporal Transformer module to extract inter-block dependencies. Specifically, the key temporal block features are flattened along the spatial dimension and treated as patches. These patches are then processed sequentially by Multi-Head Self-Attention (MSA), a Multi-Layer Perceptron (MLP), and Layer Normalization (LN), resulting in the final output representation $Z_i$ of the same dimension.

\subsection{Hierarchical Cross-Attention (HCA)}
Conventional methods typically employ simple feature concatenation or weighted averaging when modeling multi-granularity human pose features, failing to fully exploit the intrinsic relationships between features at different granularities. To fuse multi-level spatiotemporal features, we design a Hierarchical Cross-Attention (HCA) mechanism. The model extracts three levels of features $\mathbf{Z}_1, \mathbf{Z}_2, \mathbf{Z}_3$. After upsampling $\mathbf{F}_2$ and $\mathbf{Z}_3$ to the dimension of $\mathbf{Z}_1$ (denoted as $\hat{\mathbf{Z}}_2, \hat{\mathbf{Z}}_3$), the HCA module, composed of triple cross-attention units, performs the interaction as follows:
\begin{align}
\mathbf{Z}'_1 &= \text{CrossAttn}_{1}(\mathbf{Q} = \mathbf{Z}_1, \mathbf{K} = \hat{\mathbf{Z}}_2, \mathbf{V} = \hat{\mathbf{Z}}_2) \label{eq:hca1} \\
\mathbf{Z}'_2 &= \text{CrossAttn}_{2}(\mathbf{Q} = \hat{\mathbf{Z}}_2, \mathbf{K} = \mathbf{Z}_1, \mathbf{V} = \mathbf{Z}_1) \label{eq:hca2} \\
\mathbf{Z}'_3 &= \text{CrossAttn}_{3}(\mathbf{Q} = \hat{\mathbf{Z}}_3, \mathbf{K} = \mathbf{Z}_1, \mathbf{V} = \mathbf{Z}_1) \label{eq:hca3}
\end{align}
Each attention unit uses the standard multi-head attention mechanism:
\begin{equation}
\text{Attention}(\mathbf{Q}, \mathbf{K}, \mathbf{V}) = \text{Softmax}\left(\frac{\mathbf{Q}\mathbf{K}^T}{\sqrt{d_k}}\right)\mathbf{V}
\label{eq:cross_attn}
\end{equation}

\subsection{Multi-granularity Loss Function (MLF)}
The proposed multi-granularity loss combines intermediate and final representations, offering stronger supervision than traditional joint-error losses by adaptively weighting each granularity, as experimentally validated.

The network designed in this paper adopts a three-level progressive feature extraction structure. After processing human skeleton topologies of different granularities through the spatial and temporal Transformer modules, the feature tensors output at each level can be expressed as $\mathbf{Z}_1 \in \mathbb{R}^{B \times 1 \times 17 \times 16}$, $\mathbf{Z}_2 \in \mathbb{R}^{B \times 1 \times 11 \times 24}$, and $\mathbf{Z}_3 \in \mathbb{R}^{B \times 1 \times 6 \times 32}$. To fully utilize multi-granularity feature information, the intermediate features are mapped to the target space through the previously mentioned inverse-pooling operation and linear transformation, as shown in Equation~\ref{eq:loss_map}:
\begin{equation}
\mathbf{Y}_l = L_l(\text{ConvT}_{2D}([\mathbf{U}_{l}^g]^{T}, \mathbf{Z}_{l}^g)),\quad l \in \{2,3\}
\label{eq:loss_map}
\end{equation}

where $\mathbf{Z}_{l}^g$ represents the group $g$ feature of the output features $\mathbf{Z}_l$ at each level, $[\mathbf{U}_{l}^g]^T$ represents the learnable parameters of the inverse-pooling operation, $L_l(\cdot)$ represents the dimension mapping operation implemented by a fully connected layer, and $\text{ConvT}_{2D}(\cdot)$ represents the inverse-pooling operation that restores the number of joint nodes to 17.

In terms of the weight coefficient design for the hierarchical supervision strategy, this paper proposes weight settings based on clear physical significance. Since the sets of keypoints corresponding to the three-level structure of the network are $J_1=17, J_2=11, J_3=6$, the weight coefficient $\alpha_l$ is determined by the proportional relationship between the number of joint nodes at each level and the base level, as shown in Equation~\ref{eq:loss_weight}:
\begin{equation}
\alpha_l = \frac{J_l}{J_1}, \quad l \in \{1,2,3\}
\label{eq:loss_weight}
\end{equation}

From this, the weights for each layer are $\alpha_1=1, \alpha_2=11/17, \alpha_3=6/17$. The improved loss function is as follows:
\begin{equation}
\mathcal{L}_{total} = \sum_{l=1}^{3} \alpha_l \cdot \mathcal{L}_l
\label{eq:final_loss}
\end{equation}

where $\mathcal{L}_l$ is the MPJPE loss for the prediction $\mathbf{Y}_l$.

\section{Experiments}
\subsection{Datasets and Experimental Settings}
We evaluate on Human3.6M \cite{ionescu2014human36m} and HumanEva-I \cite{sigal2010humaneva}, following standard protocols from previous works \cite{pavllo2019, zheng2021poseformer, li2022mhformer}. For Human3.6M, we train on S1, S5, S6, S7, S8 and test on S9, S11. For HumanEva-I, we evaluate on three actions from S1, S2, S3. The model is implemented in PyTorch \cite{paszke2019pytorch} and trained on a single NVIDIA Tesla V100 GPU. We use the Adam optimizer \cite{kingma2014adam} with a learning rate of 0.0001, weight decay of 0.1, and stochastic depth \cite{huang2016deep} with a drop rate of 0.1. The batch size is 1024, and training runs for 200 epochs.

\subsection{Ablation Study}
We conduct ablation experiments to analyze the contribution of each module, using CPN as the 2D pose detector with an input sequence length of $T=27$. As shown in Table~\ref{tab:ablation_components}, the full model achieves an MPJPE of 46.0 mm. Removing the Multi-granularity Loss Function (MLF), Hierarchical Cross-Attention Module (HCA), Multi-granularity Spatial Transformer Module (MSTM) (include GASA and MGFE), and Multi-granularity Temporal Transformer Module (MTTM) (include TSF and KFA) progressively degrades performance, demonstrating the effectiveness of each component.

\begin{table}[htbp]
\caption{Effect of different components on model performance.}
\label{tab:ablation_components}
\centering
\begin{tabular}{cccccc}
\toprule
Model & MSTM & MTTM & HCA & MLF & MPJPE (mm) \\
\midrule
1 & $\surd$ & $\times$ & $\times$ & $\times$ & 54.9 \\
2 & $\surd$ & $\surd$ & $\times$ & $\times$ & 47.7 \\
3 & $\surd$ & $\surd$ & $\surd$ & $\times$ & 47.0 \\
4 & $\surd$ & $\surd$ & $\surd$ & $\surd$ & \textbf{46.0} \\
\bottomrule
\end{tabular}
\end{table}

We also verify the effectiveness of the MGFE and GASA mechanisms. As shown in Table~\ref{tab:ablation_mfe_gasa}, the full method achieves the lowest MPJPE of 46.0 mm. Removing either mechanism results in a performance drop, demonstrating their synergistic effect without significant computational overhead.

\begin{table}[htbp]
\caption{Effect of MFE and GASA mechanism.}
\label{tab:ablation_mfe_gasa}
\centering
\begin{tabular}{lccc}
\toprule
Method & Params (M) & FLOPs (M) & MPJPE (mm) \\
\midrule
Ours w/ MFE \& GASA & 6.39 & 43.40 & \textbf{46.0} \\
Ours w/o MFE & 6.39 & 43.36 & 47.4 \\
Ours w/o GASA & 6.39 & 43.36 & 47.5 \\
Ours w/o MFE \& GASA & 6.39 & 43.32 & 47.8 \\
\bottomrule
\end{tabular}
\end{table}

This study validates the TSF and KFA. As shown in Table~\ref{tab:ablation_tsf_csa}, their combination achieves the lowest MPJPE. Notably, removing KFA significantly increases FLOPs and degrades performance, highlighting its critical role in improving efficiency.

\begin{table}[htbp]
\caption{Effect of TSF and KFA mechanism.}
\label{tab:ablation_tsf_csa}
\centering
\begin{tabular}{lccc}
\toprule
Method & Params (M) & FLOPs (M) & MPJPE (mm) \\
\midrule
Ours w/ TSF \& KFA & 6.39 & 43.40 & \textbf{46.0} \\
Ours w/o TSF & 6.39 & 43.40 & 46.7 \\
Ours w/o KFA & 6.39 & 165.29 & 54.1 \\
Ours w/o TSF \& KFA & 6.39 & 165.29 & 55.3 \\
\bottomrule
\end{tabular}
\end{table}

Table~\ref{tab:ablation_spatial_params} presents the impact of different parameter configurations in the spatial Transformer. The configuration of 2 layers and (16,24,32) channels offers the best trade-off.

\begin{table}[htbp]
\caption{Effect of layers and feature lengths of the spatial transformer.}
\label{tab:ablation_spatial_params}
\centering
\begin{tabular}{ccccc}
\toprule
$L_S$ & $C_1, C_2, C_3$ & Params (M) & FLOPs (M) & MPJPE (mm) \\
\midrule
\multirow{3}{*}{2} & 16,16,16 & 3.99 & 26.62 & 47.3 \\
& \textbf{16,24,32} & \textbf{6.39} & \textbf{43.40} & \textbf{46.0} \\
& 16,32,48 & 9.92 & 68.71 & 46.5 \\
\midrule
\multirow{3}{*}{4} & 16,16,16 & 4.00 & 30.64 & 48.7 \\
& 16,24,32 & 6.42 & 51.03 & 47.8 \\
& 16,32,48 & 9.98 & 81.88 & 47.0 \\
\bottomrule
\end{tabular}
\end{table}

Table~\ref{tab:ablation_temporal_params} shows that for the temporal Transformer, using 4 layers and selecting $k=6$ key blocks achieves the best performance.

\begin{table}[htbp]
\caption{Effect of layers and blocks of the temporal transformer.}
\label{tab:ablation_temporal_params}
\centering
\begin{tabular}{ccccc}
\toprule
$L_T$ & $k$ & Params (M) & FLOPs (M) & MPJPE (mm) \\
\midrule
\multirow{4}{*}{2} & 4 & 3.48 & 20.19 & 49.1 \\
& 5 & 3.48 & 23.10 & 48.3 \\
& 6 & 3.48 & 26.00 & 48.1 \\
& 7 & 3.48 & 28.90 & 47.4 \\
\midrule
\multirow{4}{*}{4} & 4 & 6.39 & 31.79 & 47.8 \\
& 5 & 6.39 & 37.60 & 46.8 \\
& \textbf{6} & \textbf{6.39} & \textbf{43.40} & \textbf{46.0} \\
& 7 & 6.39 & 49.21 & 47.5 \\
\bottomrule
\end{tabular}
\end{table}

\subsection{Comparative Experiments}
Table~\ref{tab:comp_h36m} shows a comparison on Human3.6M under Protocol 1. Our method, with $T=27$ frames, achieves an average MPJPE of 46.0 mm, outperforming previous methods like Pavllo et al. \cite{pavllo2019} (48.8 mm) and Zheng et al. \cite{zheng2021poseformer} (47.0 mm).

\begin{table}[htbp]
\caption{Quantitative Comparison on Human3.6M with detected 2D poses as inputs under Protocol 1 (MPJPE in mm).}
\label{tab:comp_h36m}
\centering
\resizebox{\columnwidth}{!}{
\begin{tabular}{lccccccccccccccccc}
\toprule
Method (CPN) & Pub. & Dir. & Disc. & Eat. & Greet & Phone & Photo & Pose & Purch. & Sit & SitD. & Smoke & Wait & WalkD. & Walk & WalkT. & Avg. \\
\midrule
Pavllo et al. \cite{pavllo2019} (T=27) & CVPR'19 & - & - & - & - & - & - & - & - & 56.5 & 64.5 & 47.4 & 43.0 & 48.1 & 33.0 & 35.1 & 48.8 \\
Lin et al. \cite{lin2019trajectory} (T=50) & arXiv'19 & 42.5 & 44.8 & 42.6 & 44.2 & 48.5 & 57.1 & 52.6 & 41.4 & 56.5 & 64.5 & 47.4 & 43.0 & 48.1 & 33.0 & 35.1 & 46.6 \\
Li et al. \cite{li2022sensors} (T=27) & CVPR'20 & 43.5 & 44.8 & 43.9 & 44.1 & 47.7 & 56.5 & 44.0 & 44.2 & 55.8 & 67.9 & 47.3 & 46.5 & 45.7 & 33.4 & 33.6 & 46.6 \\
Zheng et al. \cite{zheng2021poseformer} (T=27) & ICCV'21 & - & - & - & - & - & - & - & - & - & - & - & - & - & - & - & 47.0 \\
\midrule
\textbf{Ours (T=27)} & & \textbf{42.9} & \textbf{46.8} & \textbf{47.8} & \textbf{42.9} & \textbf{46.2} & \textbf{54.4} & \textbf{47.8} & \textbf{41.8} & \textbf{47.7} & \textbf{61.7} & \textbf{47.2} & \textbf{44.5} & \textbf{52.4} & \textbf{31.6} & \textbf{35.0} & \textbf{46.0} \\
\bottomrule
\end{tabular}}
\end{table}

To evaluate generalization, we tested on HumanEva-I. As shown in Table~\ref{tab:comp_humaneva}, our method achieves an average P-MPJPE of 17.2 mm, demonstrating strong cross-dataset generalization.

\begin{table}[htbp]
\caption{Quantitative Comparison on HumanEva-I under Protocol 2 (P-MPJPE in mm).}
\label{tab:comp_humaneva}
\centering
\resizebox{\columnwidth}{!}{
\begin{tabular}{lcccccccc}
\toprule
\multirow{2}{*}{Method} & \multirow{2}{*}{Pub.} & \multicolumn{3}{c}{Walk} & \multicolumn{3}{c}{Jog} & \multirow{2}{*}{Avg.} \\
\cmidrule(lr){3-5} \cmidrule(lr){6-8}
& & S1 & S2 & S3 & S1 & S2 & S3 & \\
\midrule
Pavlakos et al. \cite{pavlakos2017coarse} & CVPR'18 & 18.8 & 12.7 & 29.2 & 23.5 & 15.4 & 14.5 & 19.0 \\
Lee et al. \cite{lee2018propagating} & ECCV'18 & 18.6 & 19.9 & 30.5 & 25.7 & 16.8 & 17.7 & 21.5 \\
Pavllo et al. \cite{pavllo2019} & CVPR'19 & 13.9 & 10.2 & 46.6 & 20.9 & 13.1 & 13.8 & 19.8 \\
Liu et al. \cite{liu2020attention} & CVPR'20 & 13.1 & 9.8 & 26.8 & 16.9 & 12.8 & 13.3 & 15.5 \\
Liu et al. \cite{liu2021graph} & ICRA'21 & 13.7 & 9.2 & 46.2 & 20.1 & 12.5 & 12.7 & 19.1 \\
Zheng et al. \cite{zheng2021poseformer} & ICCV'21 & 14.4 & 10.2 & 46.6 & 22.7 & 13.4 & 13.4 & 20.1 \\
Li et al. \cite{li2022strided} & TMM'22 & 14.0 & 10.0 & 32.8 & 19.5 & 13.6 & 14.2 & 17.4 \\
\midrule
\textbf{Ours} & & \textbf{13.2} & \textbf{12.4} & \textbf{31.3} & \textbf{19.7} & \textbf{13.1} & \textbf{13.5} & \textbf{17.2} \\
\bottomrule
\end{tabular}}
\end{table}

\begin{figure}[htbp]
\centering
\includegraphics[width=0.7\columnwidth]{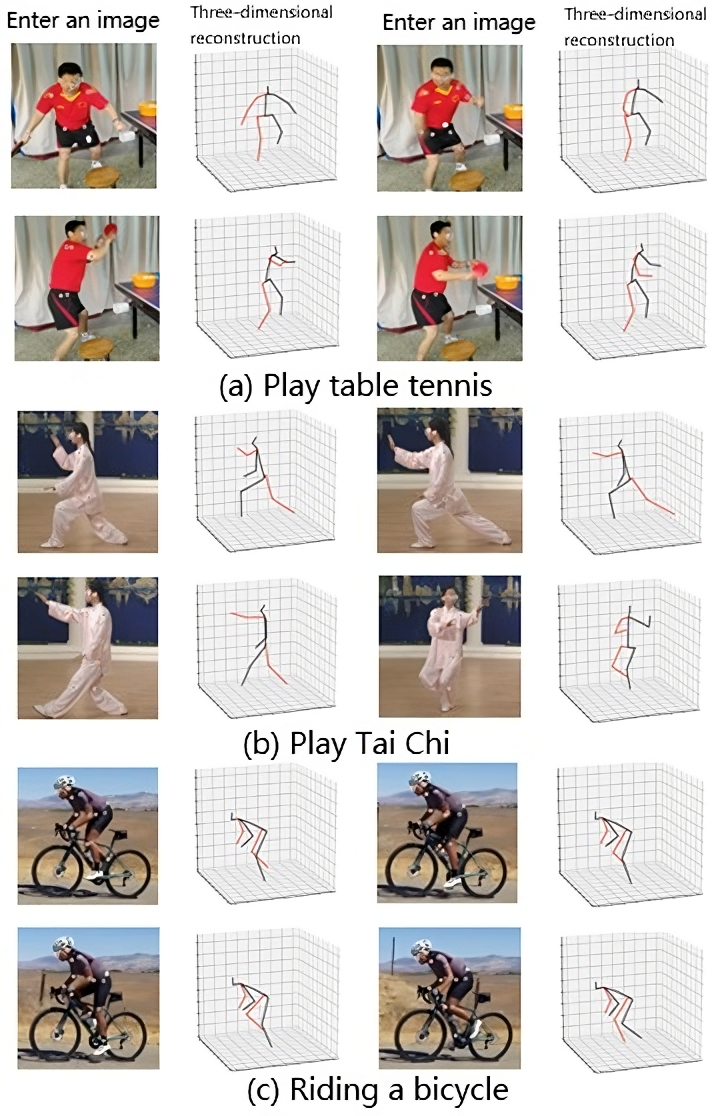}
\caption{Qualitative results of our method on in-the-wild video datasets. Our method successfully reconstructs 3D poses for various complex actions, including (a) playing table tennis, (b) practicing Tai Chi, and (c) riding a bicycle.}
\label{fig:qualitative_results}
\end{figure}

Fig.~\ref{fig:qualitative_results} illustrates qualitative results of our method on various in-the-wild videos, showcasing its robustness and practical applicability.

\section{Discussion}
This paper addresses insufficient utilization of structured information, inadequate multi-granularity feature alignment, and inefficient temporal modeling in monocular 3D HPE. On Human3.6M, our method achieves an MPJPE of 46.0 mm, outperforming PoseFormer (47.0 mm) \cite{zheng2021poseformer} and MHFormer (46.5 mm) \cite{li2022mhformer}. On HumanEva-I, our method achieves an average error of 17.2 mm, superior to competitors, validating its cross-scene generalization. Crucially, our method maintains a significantly lower computational cost (FLOPs: 43.40M) compared to MHFormer (210M+) or PoseFormer (150M+), leading to a substantial improvement in inference speed.

The MSTM, enhanced with GASA and MGFE, improves spatial awareness without increasing computational burden. The MTTM, with TSF and CSA, significantly reduces computational cost while improving accuracy by focusing on key temporal segments. Future work could explore learnable skeletal topologies, dynamic granularity selection, and cross-domain validation to further improve adaptability and generalization.

\section{Conclusion}
This paper proposed a multi-granularity spatiotemporal Transformer network that integrates structured skeletal information to address key challenges in monocular 3D HPE. Our method is designed along the lines of structure-aware modeling, multi-granularity hierarchical collaboration, and efficient keyframe selection. In spatial modeling, GASA and MGFE achieve adaptive fusion of structural priors. In temporal modeling, TSF and KFA enhance dynamic capture and efficiency. HCA facilitates collaborative feature alignment. Experimental results demonstrate that our method achieves state-of-the-art performance on public datasets like Human3.6M and HumanEva-I, significantly outperforming prior works while maintaining low computational cost. Future work will explore learnable topology optimization and dynamic granularity scheduling to further advance real-world applications.

\section{Acknowledge}
This work was supported in part by the National Natural Science Foundation of China (Grant No. 32271996).

\bibliographystyle{elsarticle-num}
\bibliography{reference}

@inproceedings{martinez2017simple,
  title={A simple yet effective baseline for 3d human pose estimation},
  author={Martinez, Julieta and Hossain, Rayat and Romero, Javier and Little, James J},
  booktitle={Proceedings of the IEEE international conference on computer vision},
  pages={2640--2649},
  year={2017}
}

@article{wang2021review,
  title={Deep 3d human pose estimation: A review},
  author={Wang, Peng and Li, Zhong-Qiu and Li, Chun-Guang and Lu, Guang},
  journal={Computers \& Visual Media},
  volume={7},
  pages={331--356},
  year={2021},
  publisher={Springer}
}

@article{sun2022survey,
  title={A survey of the state of the art in monocular 3D human pose estimation: Methods, benchmarks, and challenges},
  author={Sun, Wen and Li, Jia-Xin and Ren, Li-Zhen and Li, Chun-Guang and Lu, Guang},
  journal={IEEE Transactions on Pattern Analysis and Machine Intelligence},
  volume={45},
  number={4},
  pages={4552--4570},
  year={2022},
  publisher={IEEE}
}

@article{li2021survey,
  title={A survey on depth ambiguity of 3d human pose estimation},
  author={Li, Jia-Xin and Sun, Wen and Li, Chun-Guang and Lu, Guang},
  journal={IEEE Access},
  volume={9},
  pages={111746--111759},
  year={2021},
  publisher={IEEE}
}

@inproceedings{li2022mstformer,
  title={{MSTFormer}: Multi-granularity spatial-temporal transformers for 3d human pose estimation},
  author={Li, Lin and Liu, Xiping and Xu, Yulu and Liu, Jing and Xu, Ling},
  booktitle={European Conference on Computer Vision},
  pages={311--327},
  year={2022},
  organization={Springer}
}

@inproceedings{luo2021graformer,
  title={{GraFormer}: Graph convolution transformer for 3d pose estimation},
  author={Luo, Ziyang and Zhu, Biao and Li, Jun and Liu, Xiping and Xu, Ling},
  booktitle={Proceedings of the IEEE/CVF Conference on Computer Vision and Pattern Recognition},
  pages={12971--12980},
  year={2021}
}

@article{duan2022glagcn,
  title={{GLA-GCN}: Global-local adaptive graph convolutional network for 3D human pose estimation from monocular video},
  author={Duan, Bruce and Li, Jian},
  journal={Neurocomputing},
  volume={489},
  pages={238--249},
  year={2022},
  publisher={Elsevier}
}

@inproceedings{chen2023sct,
  title={Spectral compression transformer with line pose graph for monocular 3d human pose estimation},
  author={Chen, Wentao and Zhang, Meng-Hao and Wu, Chen},
  booktitle={Proceedings of the AAAI Conference on Artificial Intelligence},
  volume={37},
  number={1},
  pages={455--463},
  year={2023}
}

@article{li2023kinematics,
  title={Kinematics-aware spatial-temporal feature transform for 3d human pose estimation},
  author={Li, Wenkang and Liu, Hong and Li, Cheng and Lin, Chun-Yi},
  journal={Pattern Recognition},
  volume={138},
  pages={109378},
  year={2023},
  publisher={Elsevier}
}

@article{martinez2017lifting,
  title={Lifting from the deep: Convolutional 3d pose estimation from a single image},
  author={Martinez, Julieta and Raaj, Yasin and Romero, Javier and Little, James J},
  journal={arXiv preprint arXiv:1701.00295},
  year={2017}
}

@inproceedings{zhang2021unsupervised,
  title={Towards alleviating the modeling ambiguity of unsupervised monocular 3d human pose estimation},
  author={Zhang, Zixuan and Chen, Jia-Wei and Zhang, Jian-Fang},
  booktitle={Proceedings of the IEEE/CVF conference on computer vision and pattern recognition},
  pages={13589--13598},
  year={2021}
}

@inproceedings{pavllo2019,
  title={3d human pose estimation in video with temporal convolutions and semi-supervised training},
  author={Pavllo, Dario and Feichtenhofer, Christoph and Grangier, David and Auli, Michael},
  booktitle={Proceedings of the IEEE/CVF conference on computer vision and pattern recognition},
  pages={7753--7762},
  year={2019}
}

@inproceedings{chen2016exploiting,
  title={Exploiting temporal information for 3d pose estimation},
  author={Chen, Ching-Hang and Ramanan, Deva},
  booktitle={European conference on computer vision},
  pages={835--850},
  year={2016},
  organization={Springer}
}

@inproceedings{li2022liftformer,
  title={{LiftFormer}: 3d human pose estimation using attention models},
  author={Li, Ye and Shi, Jingbo and Zhang, Hong},
  booktitle={Proceedings of the IEEE/CVF conference on computer vision and pattern recognition},
  pages={11270--11279},
  year={2022}
}

@inproceedings{zheng2021poseformer,
  title={3d human pose estimation with spatial and temporal transformers},
  author={Zheng, Ce and Zhu, Sijie and Mendieta, Matias and Yang, Tao and Chen, Chen and Ding, Zhaodong},
  booktitle={Proceedings of the IEEE/CVF international conference on computer vision},
  pages={11656--11665},
  year={2021}
}

@inproceedings{zeng2022hstformer,
  title={{HSTFormer}: Hierarchical spatial-temporal transformers for 3d human pose estimation},
  author={Zeng, Wenbo and Cai, Zhaoyu and Ma, Yang and Li, Shuang},
  booktitle={Proceedings of the IEEE/CVF Conference on Computer Vision and Pattern Recognition},
  pages={11280--11289},
  year={2022}
}

@inproceedings{li2022poseformerv2,
  title={{PoseFormerV2}: Exploring frequency domain for efficient and robust 3d human pose estimation},
  author={Li, Qitao and Wang, Ce and Zhang, Yaqiao and Chen, Yujun and Wang, Licheng and Liu, Yun},
  booktitle={Proceedings of the IEEE/CVF Conference on Computer Vision and Pattern Recognition},
  pages={11259--11269},
  year={2022}
}

@article{yuan2023stie,
  title={Three-dimensional human pose estimation with spatial--temporal interaction enhancement transformer},
  author={Yuan, Jian and Chen, Hao and Zheng, Zhihao and Yu, Yue and Li, Wenbo},
  journal={Applied Sciences},
  volume={13},
  number={10},
  pages={6164},
  year={2023},
  publisher={MDPI}
}

@inproceedings{yan2019semgcn,
  title={Semantic graph convolutional networks for 3d human pose regression},
  author={Yan, Zhipeng and Peng, Ping and Sun, Xiao and Sun, Xing and He, Zongxuan},
  booktitle={Proceedings of the IEEE/CVF conference on computer vision and pattern recognition},
  pages={9007--9016},
  year={2019}
}

@inproceedings{shan2022posegtac,
  title={{PoseGTAC}: Graph transformer encoder-decoder with atrous convolution for 3d human pose estimation},
  author={Shan, Wenjing and Zhang, Ruoqing and Zhou, Yi},
  booktitle={European Conference on Computer Vision},
  pages={262--278},
  year={2022},
  organization={Springer}
}

@inproceedings{zeng2020flex,
  title={Flexible graph convolutional network for 3d human pose estimation},
  author={Zeng, Tianhan and Sun, Xiao and Sun, Xing and Liu, Wenyu},
  booktitle={European Conference on Computer Vision},
  pages={313--329},
  year={2020},
  organization={Springer}
}

@inproceedings{yan2018stgcn,
  title={Spatial temporal graph convolutional networks for skeleton-based action recognition},
  author={Yan, Sijie and Xiong, Yuanjun and Lin, Dahua},
  booktitle={Proceedings of the AAAI conference on artificial intelligence},
  volume={32},
  number={1},
  year={2018}
}

@inproceedings{ci2020posegraphnet,
  title={{PoseGraphNet++}: Enriching 3d human pose with orientation estimation},
  author={Ci, Haitao and Wang, Chunyu and Wang, Xiaokan and Sun, Yizhou},
  booktitle={European Conference on Computer Vision},
  pages={346--362},
  year={2020},
  organization={Springer}
}

@inproceedings{fang2018grammar,
  title={Learning pose grammar to encode human body configuration for 3d pose estimation},
  author={Fang, Hao-Shu and Xie, Guansong and Wang, Yuan and Lu, Cewu},
  booktitle={Proceedings of the AAAI Conference on Artificial Intelligence},
  volume={32},
  number={1},
  year={2018}
}

@article{li2022strided,
  title={Exploiting temporal contexts with strided transformer for 3d human pose estimation},
  author={Li, Wenkang and Liu, Hong and Ding, Runwei and Tang, Hong},
  journal={IEEE Transactions on Multimedia},
  volume={25},
  pages={1282--1293},
  year={2022},
  publisher={IEEE}
}

@inproceedings{cai2019gcn,
  title={Exploiting spatial-temporal relationships for 3d pose estimation via graph convolutional networks},
  author={Cai, Yujun and Ge, Liuhao and Liu, Jun and Cai, Jianfei and Cham, Tat-Jen and Yuan, Junsong and He, Ying},
  booktitle={Proceedings of the IEEE/CVF international conference on computer vision},
  pages={4052--4061},
  year={2019}
}

@inproceedings{li2022mhformer,
  title={{MHFormer}: Multi-hypothesis transformer for 3d human pose estimation},
  author={Li, Wenhao and Liu, Hong and Li, Cheng and Lin, Chun-Yi},
  booktitle={Proceedings of the IEEE/CVF conference on computer vision and pattern recognition},
  pages={13147--13156},
  year={2022}
}

@inproceedings{he2022mvg,
  title={Multiple view geometry transformers for 3d human pose estimation},
  author={He, Jianan and Yan, Jian and Yu, Yi},
  booktitle={Proceedings of the IEEE/CVF conference on computer vision and pattern recognition},
  pages={11290--11299},
  year={2022}
}

@inproceedings{zhang2022transfusion,
  title={{TransFusion}: Cross-view fusion with transformer for 3d human pose estimation},
  author={Zhang, Haoyu and Wang, Chunyu and Sun, Yizhou},
  booktitle={Proceedings of the IEEE/CVF conference on computer vision and pattern recognition},
  pages={11300--11309},
  year={2022}
}

@inproceedings{zhou2017deepkinematics,
  title={Deep kinematics analysis for monocular 3d human pose estimation},
  author={Zhou, Xingyi and Huang, Qixing and Sun, Xiao and Wei, Xiangyang and Wei, Yichen},
  booktitle={Proceedings of the IEEE conference on computer vision and pattern recognition},
  pages={2871--2880},
  year={2017}
}

@inproceedings{cibrario2024hourglass,
  title={Hourglass tokenizer for efficient transformer-based 3d human pose estimation},
  author={Cibrario, Kevin and Odobez, Jean-Marc},
  booktitle={Proceedings of the IEEE/CVF Winter Conference on Applications of Computer Vision},
  pages={5550--5559},
  year={2024}
}

@article{ionescu2014human36m,
  title={{Human3.6M}: Large scale datasets and predictive methods for 3d human sensing in natural environments},
  author={Ionescu, Catalin and Papava, Dragos and Olaru, Vlad and Sminchisescu, Cristian},
  journal={IEEE transactions on pattern analysis and machine intelligence},
  volume={36},
  number={7},
  pages={1325--1339},
  year={2013},
  publisher={IEEE}
}

@article{sigal2010humaneva,
  title={{HumanEva}: Synchronized video and motion capture dataset and baseline algorithm for evaluation of articulated human motion},
  author={Sigal, Leonid and Balan, Alexandru O and Black, Michael J},
  journal={International journal of computer vision},
  volume={87},
  pages={4--27},
  year={2010},
  publisher={Springer}
}

@inproceedings{paszke2019pytorch,
  title={Pytorch: An imperative style, high-performance deep learning library},
  author={Paszke, Adam and Gross, Sam and Massa, Francisco and Lerer, Adam and Bradbury, James and Chanan, Gregory and Killeen, Trevor and Lin, Zeming and Gimelshein, Natalia and Antiga, Luca and others},
  booktitle={Advances in neural information processing systems 32},
  year={2019}
}

@article{kingma2014adam,
  title={Adam: A method for stochastic optimization},
  author={Kingma, Diederik P and Ba, Jimmy},
  journal={arXiv preprint arXiv:1412.6980},
  year={2014}
}

@inproceedings{huang2016deep,
  title={Deep networks with stochastic depth},
  author={Huang, Gao and Sun, Yu and Liu, Zhuang and Sedra, Daniel and Weinberger, Kilian Q},
  booktitle={European conference on computer vision},
  pages={646--661},
  year={2016},
  organization={Springer}
}

@inproceedings{pavlakos2017coarse,
  title={Coarse-to-fine volumetric prediction for single-image 3d human pose},
  author={Pavlakos, Georgios and Zhou, Xingyi and Derpanis, Konstantinos G and Daniilidis, Kostas},
  booktitle={Proceedings of the IEEE conference on computer vision and pattern recognition},
  pages={7025--7034},
  year={2017}
}

@inproceedings{lee2018propagating,
  title={Propagating lstm: 3d pose estimation based on joint interdependency},
  author={Lee, Kun and Lee, In},
  booktitle={Proceedings of the European conference on computer vision (ECCV)},
  pages={119--135},
  year={2018}
}

@inproceedings{liu2020attention,
  title={Attention mechanism exploits temporal contexts: Real-time 3d human pose reconstruction},
  author={Liu, Runyang and Shen, Jiaming and Wang, He and Chen, Chen and Ding, Zhaodong},
  booktitle={Proceedings of the IEEE/CVF conference on computer vision and pattern recognition},
  pages={5064--5073},
  year={2020}
}

@inproceedings{liu2021graph,
  title={A graph attention spatio-temporal convolutional network for 3D human pose estimation in video},
  author={Liu, Jun and Rojas, Juan and Li, Yamei and Guan, Yixin},
  booktitle={2021 IEEE international conference on robotics and automation (ICRA)},
  pages={3374--3380},
  year={2021},
  organization={IEEE}
}

@article{li2022sensors,
  title={Skeleton-based Spatio-temporal U-network for 3D human pose estimation in video},
  author={Li, Wei and Du, Rui and Chen, Shangfeng},
  journal={Sensors},
  volume={22},
  number={7},
  pages={2573},
  year={2022},
  publisher={MDPI}
}

@article{lin2019trajectory,
  title={Trajectory space factorization for deep video-based 3d human pose estimation},
  author={Lin, Jintai and Lee, Gim Hee},
  journal={arXiv preprint arXiv:1908.08289},
  year={2019}
}

@inproceedings{vaswani2017attention,
  title={Attention is all you need},
  author={Vaswani, Ashish and Shazeer, Noam and Parmar, Niki and Uszkoreit, Jakob and Jones, Llion and Gomez, Aidan N and Kaiser, {\L}ukasz and Polosukhin, Illia},
  booktitle={Advances in neural information processing systems},
  pages={5998--6008},
  year={2017}
}

\end{document}